# Studies in Lower Bounding Probability of Evidence using the Markov Inequality

**Vibhav Gogate**
School of Information and Computer Science
University of California, Irvine, CA 92697
Email: vgogate@ics.uci.edu

**Bozhena Bidyuk**
Google Inc.
Irvine, CA 92612
Email: bbidyuk@google.com

**Rina Dechter**
School of Information and Computer Science
University of California, Irvine, CA 92697
Email: dechter@ics.uci.edu

**Abstract**

Computing the probability of evidence even with known error bounds is NP-hard. In this paper we address this hard problem by settling on an easier problem. We propose an approximation which provides high confidence lower bounds on probability of evidence but does not have any guarantees in terms of relative or absolute error. Our proposed approximation is a randomized importance sampling scheme that uses the Markov inequality. However, a straight-forward application of the Markov inequality may lead to poor lower bounds. We therefore propose several heuristic measures to improve its performance in practice. Empirical evaluation of our scheme with state-of-the-art lower bounding schemes reveals the promise of our approach.

## 1 Introduction

Computing the probability of evidence even with known error bounds is NP-hard [Dagum and Luby, 1993]. In this paper we address this hard problem by proposing an approximation that gives high confidence lower bounds on the probability of evidence but does not have any guarantees of relative or absolute error.

Previous work on bounding the probability of evidence comprises of *deterministic approximations* [Dechter and Rish, 2003, Leisink and Kappen, 2003, Bidyuk and Dechter, 2006a] and sampling based *randomized approximations* [Cheng, 2001, Dagum and Luby, 1997]. An approximation algorithm for computing the lower bound is deterministic if it always outputs an approximation that is a lower bound. On the other hand, an approximation algorithm is randomized if the approximation fails with some probability $\delta > 0$. The work in this paper falls under the class of randomized approximations.

Randomized approximations [Cheng, 2001, Dagum and Luby, 1997] use known inequalities such as the Chebyshev and the Hoeffding inequalities [Hoeffding, 1963] for lower (and upper) bounding the probability of evidence. The Chebyshev and Hoeffding inequalities provide bounds on how the sample mean of $N$ independently and identically distributed random variables deviates from the actual mean. The main idea in [Cheng, 2001, Dagum and Luby, 1997] is to express the problem of computing the probability of evidence as the problem of computing the mean (or expected value) of independent random variables and then use the mean over the sampled random variables to bound the deviation from the actual mean. The problem with these previous approaches is that the number of samples required to guarantee high confidence lower (or upper) bounds is inversely proportional to the probability of evidence (or the actual mean). Therefore, if the probability of evidence is arbitrarily small (e.g. $< 10^{-20}$), a large number of samples (approximately $10^{19}$) are required to guarantee the correctness of the bounds.

We alleviate this problem, which arises from the dependence of the Hoeffding and Chebyshev inequalities on the number of samples $N$, by using the Markov inequality which is independent of $N$. Recently, the Markov inequality was used to lower bound the number of solutions of a Satisfiability formula [Gomes et al., 2007] showing good empirical results. We adapt this scheme to compute lower bounds on probability of evidence and extend it in several ways. First, we show how importance sampling can be used to obtain lower bounds using the Markov inequality. Second, we address the difficulty associated with the approach presented in [Gomes et al., 2007] in that with the increase in number of samples the lower bound is likely to decrease by proposing several parametric heuristic methods. Third, we show how the probability of evidence of belief networks with zero probabilities can be efficiently estimated by using the Markov inequality in conjunction with a recently proposed SampleSearch scheme [Gogate and Dechter, 2007]. Finally, we provide empirical results demonstrating the potential of our new scheme by comparing against state-of-the-art bounding schemes such as bound propagation [Leisink and Kappen, 2003] and its improvements [Bidyuk and Dechter, 2006b].

The rest of the paper is organized as follows. In section 2, we discuss preliminaries and related work. In section 3, we present our lower bounding scheme and propose various heuristics to improve it. In section 4, we describe how the SampleSearch scheme can be used within our lower bounding scheme. Experimental results are presented in section 5 and we end with a summary in section 6.

## 2 Preliminaries and Previous work

We represent sets by bold capital letters and members of a set by capital letters. An assignment of a value to a variable is denoted by a small letter while bold small letters indicate an assignment to a set of variables.

**Definition 1. (belief networks)** A *belief network (BN)* is a graphical model $\mathscr{P} = \langle \mathbf{Z}, \mathbf{D}, \mathbf{P} \rangle$, where $\mathbf{Z} = \{Z_1, \ldots, Z_n\}$ is a set of random variables over multi-valued domains $\mathbf{D} = \{\mathbf{D_1}, \ldots, \mathbf{D_n}\}$. Given a directed acyclic graph $G$ over $\mathbf{Z}$, $\mathbf{P} = \{P_i\}$, where $P_i = P(Z_i|\mathbf{pa}(Z_i))$ are conditional probability tables (CPTs) associated with each $Z_i$. The set $\mathbf{pa}(Z_i)$ is the set of parents of the variable $Z_i$ in $G$. A belief network represents a probability distribution over $\mathbf{Z}$, $P(\mathbf{Z}) = \prod_{i=1}^{n} P(Z_i|\mathbf{pa}(Z_i))$. An *evidence set* $\mathbf{E} = \mathbf{e}$ is an instantiated subset of variables.

**Definition 2** (Probability of Evidence). Given a belief network $\mathscr{P}$ and evidence $\mathbf{E} = \mathbf{e}$, the probability of evidence $P(\mathbf{E} = \mathbf{e})$ is defined as:

$$P(\mathbf{e}) = \sum_{\mathbf{Z} \setminus \mathbf{E}} \prod_{j=1}^{n} P(Z_j|\mathbf{pa}(Z_j))_{|\mathbf{E}=\mathbf{e}} \quad (1)$$

The notation $h(\mathbf{Z})_{|\mathbf{E}=\mathbf{e}}$ stands for a function $h$ over $\mathbf{Z} \setminus \mathbf{E}$ with the assignment $\mathbf{E} = \mathbf{e}$.

### 2.1 Computing Probability of Evidence Using Importance Sampling

Importance sampling [Rubinstein, 1981] is a simulation technique commonly used to evaluate the following sum: $M = \sum_{\mathbf{x} \in \mathbf{X}} f(\mathbf{x})$ for some real function $f$. The idea is to generate samples $\mathbf{x}^1, \ldots, \mathbf{x}^N$ from a proposal distribution $Q$ (satisfying $f(\mathbf{x}) > 0 \Rightarrow Q(\mathbf{x}) > 0$) and then estimate $M$ as follows:

$$M = \sum_{\mathbf{x} \in \mathbf{X}} f(\mathbf{x}) = \sum_{\mathbf{x} \in \mathbf{X}} \frac{f(\mathbf{x})}{Q(\mathbf{x})} Q(\mathbf{x}) = \mathbb{E}_Q[\frac{f(\mathbf{x})}{Q(\mathbf{x})}] \quad (2)$$

$$\widehat{M} = \frac{1}{N} \sum_{i=1}^{N} w(\mathbf{x}^i) , \text{ where } w(\mathbf{x}^i) = \frac{f(\mathbf{x}^i)}{Q(\mathbf{x}^i)} \quad (3)$$

$w$ is often referred to as the sample weight. It is known that the expected value $\mathbb{E}(\widehat{M}) = M$ [Rubinstein, 1981].

To compute the probability of evidence by importance sampling, we use the substitution:

$$f(\mathbf{x}) = P(\mathbf{z}, \mathbf{e}) = \prod_{j=1}^{n} P(Z_j|\mathbf{pa}(Z_j))_{|\mathbf{E}=\mathbf{e}} , \ \mathbf{X} = \mathbf{Z} \setminus \mathbf{E} \quad (4)$$

For the rest of the paper, assume $M = P(\mathbf{e})$ and $f(\mathbf{x}) = \prod_{j=1}^{n} P(Z_j|\mathbf{pa}(Z_j))_{|\mathbf{E}=\mathbf{e}}$.

Several choices are available for the proposal distribution $Q(\mathbf{x})$ ranging from the prior distribution as in likelihood weighting to more sophisticated alternatives such as IJGP-Sampling [Gogate and Dechter, 2005] and EPIS-BN [Yuan and Druzdzel, 2006] where the output of belief propagation is used to compute the proposal distribution.

As in prior work [Cheng and Druzdzel, 2000], we assume that the *proposal distribution is expressed in a factored product form* dictated by the belief network: $Q(\mathbf{X}) = \prod_{i=1}^{n} Q_i(X_i|X_1, \ldots, X_{i-1}) = \prod_{i=1}^{n} Q_i(X_i|\mathbf{Y_i})$, where $\mathbf{Y_i} \subseteq \{X_1, \ldots, X_{i-1}\}$, $Q_i(X_i|\mathbf{Y_i}) = Q(X_i|X_1, \ldots, X_{i-1})$ and $|\mathbf{Y_i}| < c$ for some constant $c$. When $Q$ is given in a product form, we can generate a full sample from $Q$ as follows. For $i = 1$ to $n$, sample $X_i = x_i$ from the conditional distribution $Q(X_i|X_1 = x_1, \ldots, X_{i-1} = x_{i-1})$ and set $X_i = x_i$. This is often referred to as an *ordered Monte Carlo sampler*.

### 2.2 Related Work

[Dagum and Luby, 1997] provide an upper bound on the number of samples $N$ required to guarantee that for any $\varepsilon, \delta > 0$, the estimate $\widehat{M}$ computed using Equation 3 approximates $M$ with relative error $\varepsilon$ with probability at least $1 - \delta$. Formally,

$$\mathbf{Pr}[M(1-\varepsilon) \leq \widehat{M} \leq M(1+\varepsilon)] > 1 - \delta \quad (5)$$

The specific bound on $N$ that the authors derive is:

$$N \geq \frac{4}{M\varepsilon^2} ln \frac{2}{\delta} \quad (6)$$

These bounds were later improved by [Cheng, 2001] to yield:

$$N \geq \frac{1}{M} \frac{1}{(1+\varepsilon)ln(1+\varepsilon) - \varepsilon} ln \frac{2}{\delta} \quad (7)$$

In both these bounds (Equations 6 and 7) $N$ is inversely proportional to $M$ and therefore when $M$ is small, a large number of samples are required to achieve an acceptable confidence level $(1 - \delta) > 99\%$.

A bound on $N$ is required because [Dagum and Luby, 1997, Cheng, 2001] use Chebyshev and Hoeffding inequalities which depend on $N$ for correctness. Instead, we could use the Markov inequality which is independent of $N$ and still achieve high confidence lower bounds. The independence from $N$ allows us to use even a single sample to derive lower bounds. The only caveat is that our proposed method does not have any guarantees in terms of relative error $\varepsilon$. We describe our method in the next section.

## 3 Markov Inequality to lower bound P(e)

**Definition 3** (Markov Inequality). For any random variable $X$ and $k > 1$, $Pr(X \geq k\mathbb{E}[X]) \leq \frac{1}{k}$

[Gomes et al., 2007] show how the Markov inequality can be used to obtain probabilistic lower bounds on the number of solutions of a satisfiability/constraint satisfaction problem. Using the same approach, we present a small modification of importance sampling for obtaining lower bounds on the probability of evidence (see Algorithm 1). The algorithm generates $k$ independent samples from a proposal distribution $Q$ and returns the minimum $\frac{f(\mathbf{x})}{\alpha Q(\mathbf{x})}$ (minCount in Algorithm 1) over the $k$-samples.

---

**Algorithm 1** Markov-LB $(f, Q, k, \alpha > 1)$

1: $minCount \leftarrow \infty$
2: **for** $i = 1$ to $k$ **do**
3:   Generate a sample $\mathbf{x}^i$ from $Q$
4:   **IF** $minCount > \frac{f(\mathbf{x}^i)}{\alpha * Q(\mathbf{x}^i)}$ **THEN** $minCount = \frac{f(\mathbf{x}^i)}{\alpha * Q(\mathbf{x}^i)}$
5: **end for**
6: **Return** $minCount$

---

THEOREM 1 (Lower Bound). *With probability of at least $1 - 1/\alpha^k$, Markov-LB returns a lower bound on $M = P(\mathbf{e})$*

*Proof.* Consider an arbitrary sample $\mathbf{x}^i$. It is clear from the discussion in section 2 that the expected value of $f(\mathbf{x}^i)/Q(\mathbf{x}^i)$ is equal to $M$. Therefore, by the Markov inequality, we have $\mathbf{Pr}(\frac{f(\mathbf{x}^i)}{\alpha * Q(\mathbf{x}^i)} > M) < 1/\alpha$. Since, the generated $k$ samples are independent, the probability $\mathbf{Pr}(min_{i=1}^k \frac{f(\mathbf{x}^i)}{\alpha * Q(\mathbf{x}^i)} > M) < 1/\alpha^k$ and therefore $\mathbf{Pr}(min_{i=1}^k [\frac{f(\mathbf{x}^i)}{\alpha * Q(\mathbf{x}^i)}] \leq M) > 1 - \frac{1}{\alpha^k}$. □

The problem with Algorithm 1 is that unless the variance of $f(\mathbf{x})/Q(\mathbf{x})$ is very small, we expect the lower bound to decrease with increase in the number of samples $k$. In practice, given a required confidence of $\delta = \alpha^k$, one can decrease $\alpha$ as $k$ is increased.

Note that each sample in Algorithm 1 provides a lower bound with probability $> (1 - 1/\alpha)$. We can replace the sample by any procedure that provides a lower bound with probability $> (1 - 1/\alpha)$ and therefore in the following we propose several heuristic methods to compute a lower bound with probability $> (1 - 1/\alpha)$.

### 3.1 Using Average over $N$ samples

One obvious way is to use the importance sampling estimator $\widehat{M}$. Because $\mathbb{E}[\widehat{M}] = M$, by Markov inequality $\widehat{M}/\alpha$ is a lower bound of $M$ with confidence $1 - 1/\alpha$. As the number of samples increases, the average becomes more stable and is likely to increase the minimum value over the $k$ iterations of Algorithm 1.

### 3.2 Using the maximum over $N$ samples

We can even use the maximum instead of the average over the $N$ i.i.d samples and still achieve a confidence of $1 - 1/\alpha$. Given a set of $N$ independent events such that each event occurs with probability $> (1 - 1/\beta)$, the probability that all events occur is $> (1 - 1/\beta)^N$. Consequently, we can prove that:

**Proposition 1.** *Given $N$ i.i.d. samples generated from $Q$, $\mathbf{Pr}(max_{i=1}^N [\frac{f(\mathbf{x}^i)}{\beta Q(\mathbf{x}^i)}] < M) > (1 - 1/\beta)^N$.*

Therefore, by setting $(1 - 1/\beta)^N = 1 - 1/\alpha$ (i.e. $\beta = 1/[1 - (1 - 1/\alpha)^{1/N}]$) and recording the maximum value of $f(\mathbf{x}^i)/\beta Q(\mathbf{x}^i)$ over the $N$ samples, we can achieve a lower bound on $M$ with confidence $(1 - 1/\alpha)$.

Again the problem with this method is that increasing the number of samples increases $\beta$ and consequently the lower bound decreases. However, when the variance of the random variables $f(\mathbf{x}^i)/Q(\mathbf{x}^i)$ is large, the maximum value is likely to be larger than the sample average. Another approach to utilize the maximum over the $N$ samples is to use the martingale inequalities.

### 3.3 Using the martingale Inequalities

In this subsection, we show how the martingale theory can be used to obtain lower bounds on $P(\mathbf{e})$.

**Definition 4** (Martingale). A sequence of random variables $X_1, \ldots, X_N$ is a martingale with respect to another sequence $Z_1, \ldots, Z_N$ defined on a common probability space $\Omega$ iff $\mathbb{E}[X_i | Z_1, \ldots, Z_{i-1}] = X_{i-1}$ for all $i$.

Given i.i.d. samples $(\mathbf{x}^1, \ldots, \mathbf{x}^N)$ generated from $Q$, note that the sequence $\Lambda_1, \ldots, \Lambda_N$, where $\Lambda_p = \prod_{i=1}^p \frac{f(\mathbf{x}^i)}{MQ(\mathbf{x}^i)}$ forms a martingale as shown below:

$$\mathbb{E}[\Lambda_p | \mathbf{x}^1, \ldots, \mathbf{x}^{p-1}] = \mathbb{E}[\Lambda_{p-1} * \frac{f(\mathbf{x}^p)}{M * Q(\mathbf{x}^p)} | \mathbf{x}^1, \ldots, \mathbf{x}^{p-1}]$$
$$= \Lambda_{p-1} * \mathbb{E}[\frac{f(\mathbf{x}^p)}{M * Q(\mathbf{x}^p)} | \mathbf{x}^1, \ldots, \mathbf{x}^{p-1}]$$

Because, $\mathbb{E}[\frac{f(\mathbf{x}^p)}{M*Q(\mathbf{x}^p)} | \mathbf{x}^1, \ldots, \mathbf{x}^{p-1}] = 1$, we have $\mathbb{E}[\Lambda_p | \mathbf{x}^1, \ldots, \mathbf{x}^{p-1}] = \Lambda_{p-1}$ as required. The expected value $\mathbb{E}[\Lambda_1] = 1$ and for such martingales which have a mean of 1, [Breiman, 1968] provides the following extension of the Markov inequality:

$$\mathbf{Pr}(max_{i=1}^N \Lambda_i > \alpha) \leq \frac{1}{\alpha} \quad (8)$$

and therefore,

$$\mathbf{Pr}((max_{i=1}^N \prod_{j=1}^i \frac{f(\mathbf{x}^j)}{MQ(\mathbf{x}^j)}) > \alpha) \leq \frac{1}{\alpha} \quad (9)$$

From Inequality 9, we can see that $max_{i=1}^N (\frac{1}{\alpha} \prod_{j=1}^i [\frac{f(\mathbf{x}^j)}{Q(\mathbf{x}^j)}])^{1/i}$ is a lower bound on $M$ with a confidence of $(1-1/\alpha)$. In general one could use any randomly selected permutation of the samples $(\mathbf{x}^1,\ldots,\mathbf{x}^N)$ and apply inequality 9.

Another related extension of Markov inequality for martingales deals with the order statistics of the sample. Let $\frac{f(\mathbf{x}^{(1)})}{MQ(\mathbf{x}^{(1)})} \leq \frac{f(\mathbf{x}^{(2)})}{MQ(\mathbf{x}^{(2)})} \leq \ldots \leq \frac{f(\mathbf{x}^{(N)})}{MQ(\mathbf{x}^{(N)})}$ be the order statistics of the sample. Using martingale theory, [Kaplan, 1987] proved that the random variable

$$\Theta^* = max_{i=1}^N \prod_{j=1}^i \frac{f(\mathbf{x}^{(N-j+1)})}{M*Q(\mathbf{x}^{(N-j+1)})*\binom{N}{i}}$$

satisfies the inequality $\mathbf{Pr}(\Theta^* > \alpha) \leq 1/\alpha$. Therefore,

$$\mathbf{Pr}((max_{i=1}^N \prod_{j=1}^i \frac{f(\mathbf{x}^{(N-j+1)})}{M*Q(\mathbf{x}^{(N-j+1)})*\binom{N}{i}}) > \alpha) \leq \frac{1}{\alpha} \quad (10)$$

From Inequality 10, we can see that $max_{i=1}^N (\frac{1}{\alpha} \prod_{j=1}^i [\frac{f(\mathbf{x}^{(N-j+1)})}{Q(\mathbf{x}^{(N-j+1)})\binom{N}{i}}])^{1/i}$ is a lower bound on $M$ with a confidence of $(1-1/\alpha)$.

To summarize in this section, we have proposed four heuristic ways to improve Algorithm 1 (1) The average method, (2) The max method and (3) The martingale random permutation method and (4) The martingale order statistics method.

## 4 Overcoming Rejection: Using SampleSearch with Markov-LB

One problem with importance sampling based algorithms is the so-called *rejection problem* and in this section we discuss how to alleviate this problem in Markov-LB by using the recently proposed SampleSearch scheme [Gogate and Dechter, 2007].

### 4.1 Rejection Problem

Given a positive belief network that expresses the probability distribution $\mathscr{P}(\mathbf{Z}) = \prod_{i=1}^n P(Z_i|Z_1,\ldots,Z_{i-1})$ and an empty evidence set, all full samples generated by the ordered Monte Carlo sampler along the ordering $Z_1,\ldots,Z_n$ are guaranteed to have a non-zero weight. However, in presence of both zero probabilities and evidence the ordered Monte Carlo sampler may generate samples which have zero weight because the sample may conflict with the evidence and zero probabilities. Formally, if the proposal distribution $Q$ is such that the probability of sampling an assignment from the set $\{\mathbf{x}|f(\mathbf{x})=0\}$ is substantially larger than the probability of sampling an assignment from the set $\{\mathbf{x}|f(\mathbf{x})>0\}$, a large number of samples generated from $Q$ will have zero weight. In fact, in the extreme case if no positive weight samples are generated, the lower bound reported by the Markov-LB scheme will be trivially zero.

The rejection problem has been largely ignored in the importance sampling community except the work on adaptive importance sampling techniques [Hernández et al., 1998, Cheng and Druzdzel, 2000, Yuan and Druzdzel, 2006]. In [Gogate and Dechter, 2005], we initiated a new approach of reducing the amount of rejection by using constraint processing methods. The main idea is to express the zero probabilities in the belief network using constraints.

**Definition 5** (constraint network). A constraint network (CN) is defined by a 3-tuple, $\mathscr{R} = \langle \mathbf{Z}, \mathbf{D}, \mathbf{C} \rangle$, where $\mathbf{Z}$ is a set of variables $\mathbf{Z} = \{Z_1,\ldots,Z_n\}$, associated with a set of discrete-valued domains, $\mathbf{D} = \{\mathbf{D_1},\ldots,\mathbf{D_n}\}$, and a set of constraints $\mathbf{C} = \{C_1,\ldots,C_r\}$. Each constraint $C_i$ is a relation $\mathbf{R_{S_i}}$ defined on a subset of variables $\mathbf{S_i} \subseteq \mathbf{Z}$. The relation denotes all compatible tuples of the cartesian product of the domains of $\mathbf{S_i}$. A solution is an assignment of values to all variables $\mathbf{z} = (Z_1 = z_1,\ldots,Z_n = z_n)$, $z_i \in \mathbf{D}_i$, such that $\mathbf{z}$ belongs to the natural join of all constraints i.e. $\mathbf{z} \in \mathbf{R_{S_1}} \bowtie \ldots \bowtie \mathbf{R_{S_r}}$. The constraint satisfaction problem (CSP) is to determine if a constraint network has a solution, and if so, to find one. When we write $\mathscr{R}(\mathbf{z})$, we mean that $\mathbf{z}$ satisfies all the constraints in $\mathscr{R}$.

In the following example, we show how constraints can be extracted from the CPTs.

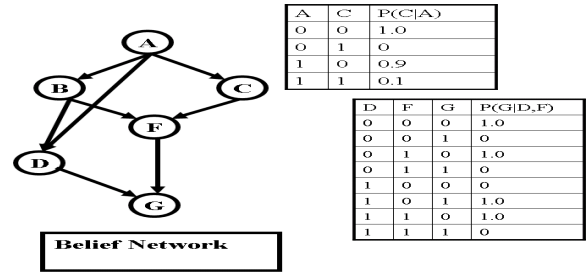

Figure 1: An example Belief Network.

**Example 1.** Figure 1 presents a belief network over 6 binary variables. The CPTs associated with $C$ and $G$ have zero probabilities. The constraint extracted from the CPT of $C$ is the relation $R_{A,C} = \{(0,0),(1,0),(1,1)\}$ while the CPT of $G$ yields the constraint relation $R_{D,F,G} = \{(0,0,0)(0,1,0),(1,0,1),(1,1,0)\}$. Namely, each "0" tuple in a CPT corresponds to a no-good, and therefore does not appear in the corresponding relation.

Our importance sampling scheme called IJGP-Sampling [Gogate and Dechter, 2005] uses constraint propagation to reduce rejection. Given a partial sample $(x_1,\ldots,x_p)$, constraint propagation prunes values in the domains of future variables $X_{p+1},\ldots,X_n$ which are inconsistent with $(x_1,\ldots,x_p)$.

However, we observed recently that when a substantial number of zero probabilities are present or when there are many evidence variables, the level of constraint propagation achieved by IJGP is not effective and often few/no consistent samples will be generated. Therefore in [Gogate and Dechter, 2007], we proposed a more aggressive approach that searches explicitly for a non-zero weight sample yielding the *SampleSearch* scheme.

---

**Algorithm 2** SampleSearch
**Input:** The proposal distribution $Q(\mathbf{x}) = \prod_{i=1}^{n} Q_i(x_i|x_1,\ldots,x_{i-1})$, hard constraints $\mathscr{R}$ that represent zeros in $f(\mathbf{x})$
**Output:** A sample $\mathbf{x} = (x_1,\ldots,x_n)$ satisfying all constraints in $\mathscr{R}$

---
1: $i = 1$, $D'_i = D_i$ (copy domains), $Q'_i(X_i) = Q_i(X_i)$ (copy distribution), $\mathbf{x} = \phi$
2: **while** $1 \leq i \leq n$ **do**
3:   **if** $D'_i$ is not empty **then**
4:     Sample $X_i = x_i$ from $Q'_i$ and remove it from $D'_i$
5:     **if** $(x_1,\ldots,x_i)$ violates any constraints in $\mathscr{R}$ **then**
6:       set $Q_i(x_i|x_1,\ldots,x_{i-1}) = 0$ and normalize $Q'_i$
7:       **Goto** Step 3
8:     **end if**
9:     $\mathbf{x} = \mathbf{x} \cup x_i$, $i = i+1$, $D'_i = D_i$, $Q'_i(X_i|x_1,\ldots,x_{i-1}) = Q_i(X_i|x_1,\ldots,x_{i-1})$
10:   **else**
11:     $\mathbf{x} = \mathbf{x} \setminus x_{i-1}$.
12:     set $Q'_{i-1}(X_{i-1} = x_{i-1}|x_1,\ldots,x_{i-2}) = 0$ and normalize.
13:     set $i = i - 1$
14:   **end if**
15: **end while**
16: **Return x**

---

### 4.2 The SampleSearch scheme

An ordered Monte Carlo sampler samples variables in the order $\langle X_1,\ldots,X_n \rangle$ from the proposal distribution $Q$ and rejects a partial or full sample $(x_1,\ldots,x_i)$ if it violates any constraints in $\mathscr{R}$ ($\mathscr{R}$ models zero probabilities in $f$). Upon rejecting a (partial or full) sample, the sampler starts sampling anew from the first variable in the ordering. Instead, we propose the following modification. We can set $Q_i(X_i = x_i|x_1,\ldots,x_{i-1}) = 0$ (to reflect that $(x_1,\ldots,x_i)$ is not consistent), normalize $Q_i$ and re-sample $X_i$ from the normalized distribution. The newly sampled value may be consistent in which case we can proceed to variable $X_{i+1}$ or it may be inconsistent. If we repeat the process we may reach a point where $Q_i(X_i|x_1,\ldots,x_{i-1})$ is 0 for all values of $X_i$. In this case, $(x_1,\ldots,x_{i-1})$ is inconsistent and therefore we need to change the distribution at $X_{i-1}$ by setting $Q_{i-1}(X_{i-1} = x_{i-1}|x_1,\ldots,x_{i-2}) = 0$, normalize and re-sample $X_{i-1}$. We can repeat this process until a globally consistent full sample that satisfies all constraints in $\mathscr{R}$ is generated. By construction, this process always yields a globally consistent full sample.

Our proposed SampleSearch scheme is described as Algorithm 2. It is a depth first backtracking search (DFS) over the state space of consistent partial assignments searching for a solution to a constraint satisfaction problem $\mathscr{R}$, whose value selection is guided by $Q$.

It can be proved that SampleSearch generates independently and identically distributed samples from the backtrack-free distribution which we define below.

**Definition 6** (Backtrack-free distribution ). Given a distribution $Q(\mathbf{X}) = \prod_{i=1}^{N} Q_i(X_i|X_1,\ldots,X_{i-1})$, an ordering $O = \langle x_1,\ldots,x_n \rangle$ and a set of constraints $\mathscr{R}$, the backtrack-free distribution $Q^R$ is the distribution:

$$Q^R(\mathbf{x}) = \prod_{i=1}^{n} Q_i^R(x_i|x_1,\ldots,x_{i-1}) \quad (11)$$

where $Q_i^R(x_i|x_1,\ldots,x_{i-1})$ is given by:

$$Q_i^R(x_i|x_1,\ldots,x_{i-1}) = \frac{Q_i(x_i|x_1,\ldots,x_{i-1})}{1 - \sum_{x'_i \in \mathbf{B_i}} Q_i(x'_i|x_1,\ldots,x_{i-1})} \quad (12)$$

where $\mathbf{B_i} = \{x'_i \in \mathbf{D_i}|(x_1,\ldots,x_{i-1},x'_i)$ can not be extended to a solution of $\mathscr{R}\}$ and $x_i \notin \mathbf{B_i}$. Note that by definition, $f(\mathbf{x}) = 0 \Rightarrow Q^R(\mathbf{x}) = 0$ and vice versa.

THEOREM 2. *[Gogate and Dechter, 2007] SampleSearch generates independently and identically distributed samples from the backtrack-free distribution.*

Given that the backtrack-free distribution is the sampling distribution of SampleSearch, we can use SampleSearch within the importance sampling framework as follows. Let $(\mathbf{x}^1,\ldots,\mathbf{x}^N)$ be a set of i.i.d samples generated by SampleSearch. Then we can estimate $M = \sum_{\mathbf{x} \in \mathbf{X}} f(\mathbf{x})$ as:

$$\widehat{M} = \frac{1}{N} \sum_{i=1}^{N} \frac{f(\mathbf{x}^i)}{Q^R(\mathbf{x}^i)} \quad (13)$$

Although SampleSearch was described using the naive backtracking algorithm, in principle we can integrate any systematic CSP/SAT solver that employs advanced search schemes with sampling through our SampleSearch scheme. Since the current implementations of SAT solvers are very efficient, we represent the zero probabilities in the belief network using cnf (SAT) expressions and use Minisat [Sorensson and Een, 2005] as our SAT solver.

**Computing $Q^R(\mathbf{x})$ given a sample x**:
From Definition 6, we notice that to compute the components $Q_i^R(x_i|x_1,\ldots,x_{i-1})$ for a sample $\mathbf{x} = (x_1,\ldots,x_n)$, we have to determine the set $\mathbf{B_i} = \{x'_i \in \mathbf{D_i}|(x_1,\ldots,x_{i-1},x'_i)$ can not be extended to a solution of $\mathscr{R}\}$. The set $\mathbf{B_i}$ can be determined by checking for each $x'_i \in \mathbf{D_i}$ if the partial assignment $(x_1,\ldots,x_{i-1},x'_i)$ can be extended to a solution of $\mathscr{R}$. To speed up this checking at each branch point, we use the Minisat SAT solver [Sorensson and Een, 2005]. Minisat should be invoked a maximum of $O(n * (d-1))$ times where $n$ is the number of variables and $d$ is the maximum domain size.

In [Gogate and Dechter, 2007], we found that the SampleSearch based importance sampling scheme outperforms all competing approaches when a substantial number of zero probabilities are present in the belief network. Therefore, we employ SampleSearch as a sampling technique within Markov-LB when a substantial number of zero probabilities are present. It should be obvious that when SampleSearch is used, we should use $\frac{f(\mathbf{x}^i)}{Q^R(\mathbf{x}^i)}$ as a random variable instead of $\frac{f(\mathbf{x}^i)}{Q(\mathbf{x}^i)}$ in the Markov-LB scheme.

## 5 Empirical Evaluation

### 5.1 Competing Algorithms

**Markov-LB with SampleSearch and IJGP-sampling:** The performance of importance sampling based algorithms is highly dependent on the proposal distribution [Cheng and Druzdzel, 2000, Yuan and Druzdzel, 2006]. It was shown that computing the proposal distribution from the output of a Generalized Belief Propagation scheme of Iterative Join Graph Propagation (IJGP) yields better empirical performance than other available choices [Gogate and Dechter, 2005]. Therefore, we use the output of IJGP to compute the proposal distribution $Q$. The complexity of IJGP is time and space exponential in its $i$-bound, a parameter that bounds cluster sizes. We use a $i$-bound of 3 in all our experiments. The preprocessing time for computing the proposal distribution using IJGP ($i = 3$) was negligible ($< 2$ seconds for the hardest instances).

We experimented with four versions of Markov-LB (a) Markov-LB as given in Algorithm 1, (b) Markov-LB with the average heuristic, (c) Markov-LB with the martingale random permutation heuristic and (d) Markov-LB with the martingale order statistics heuristic. In all our experiments, we set $\alpha = 2$ and $k = 7$ which gives us a correctness confidence of $1 - 1/2^7 \approx 99.2\%$ on our lower bounds. Finally, we set $N = 100$ for the heuristic methods. Also note that when the belief network is positive we use IJGP-sampling but when the belief network has zero probabilities, we use SampleSearch whose initial proposal distribution $Q$ is computed from the output of IJGP.

**Bound Propagation with Cut-set Conditioning** We also experimented with the state of the art any-time bounding scheme that combines sampling-based cut-set conditioning and bound propagation [Leisink and Kappen, 2003] and which is a part of Any-Time Bounds framework for bounding posterior marginals [Bidyuk and Dechter, 2006a]. Given a subset of variables $\mathbf{C} \subset \mathbf{X} \setminus \mathbf{E}$, we can compute $P(\mathbf{e})$ exactly as follows:

$$P(\mathbf{e}) = \sum_{i=1}^{k} P(\mathbf{c}^\mathbf{i}, \mathbf{e}) \quad (14)$$

The lower bound on $P(\mathbf{e})$ is obtained by computing $P(\mathbf{c}^\mathbf{i}, \mathbf{e})$ for $h$ high probability tuples of $\mathbf{C}$ (selected through sampling) and bounding the remaining probability mass by computing a lower bound $P^L(\mathbf{c_1}, ..., \mathbf{c_q}, \mathbf{e})$ on $P(\mathbf{c_1}, ..., \mathbf{c_q}, \mathbf{e})$, $q < |\mathbf{C}|$, for a polynomial number of partially instantiated tuples of subset $C$, resulting in:

$$P(\mathbf{e}) \geq \sum_{i=1}^{h} P(\mathbf{c^i}, \mathbf{e}) + \sum_{i=1}^{k'} P_{BP}^L(\mathbf{c_1^i}, ..., \mathbf{c_q^i}, \mathbf{e}) \quad (15)$$

where lower bound $P_{BP}^L(\mathbf{c_1}, ..., \mathbf{c_q}, \mathbf{e})$ is obtained using bound propagation. Although bound propagation bounds marginal probabilities, it can be used to bound any joint probability $P(\mathbf{z})$ as follows:

$$P_{BP}^L(\mathbf{z}) = \prod_i P_{BP}^L(z_i | z_1, ..., z_{i-1}) \quad (16)$$

where lower bound $P_{BP}^L(z_i|z_1,...,z_{i-1})$ is computed directly by bound propagation. We use here the same variant of bound propagation described in [Bidyuk and Dechter, 2006b] that is used by the Any-Time Bounds framework. The lower bound obtained by Eq. 15 can be improved by exploring a larger number of tuples $h$. After generating $h$ tuples by sampling, we can stop the computation at any time after bounding $p < k'$ out of $k'$ partially instantiated tuples and produce the result.

In our experiments we run the bound propagation with cut-set conditioning scheme until convergence or until a stipulated time bound has expired. Finally, we should note that the bound propagation with cut-set conditioning scheme provides deterministic lower and upper bounds on $P(\mathbf{e})$ while our Markov-LB scheme provides only a lower bound and it may fail with a probability $\delta \leq 0.01$.

### 5.1.1 Evaluation Criteria

We experimented with six sets of benchmark belief networks (a) Alarm networks (b) CPCS networks, (c) Randomly generated belief networks, (d) Linkage networks, (e) Grid networks and (f) Two-layered deterministic networks. Note that only linkage, grid and deterministic networks have zero probabilities.

On each network instance, we compare log relative error between the exact probability of evidence and the lower bound reported by the competing techniques. Formally, if $P_{exact}$ is the actual probability of evidence and $P_{app}$ is the approximate probability of evidence, we compute the log-relative error as follows:

$$\Delta = Abs(\frac{log(P_{exact}) - log(P_{app})}{log(P_{exact})}) \quad (17)$$

Note that the exact $P(\mathbf{e})$ for most instances is available from the UAI competition web-site [1]. The exact $P(\mathbf{e})$ for the two layered deterministic networks was computed using AND/OR search [Dechter and Mateescu, 2004].

We compute the log relative error instead of the usual relative error because when the probability of evidence is extremely small the relative error between the exact and the approximate probability of evidence will be arbitrarily close to 1 and we would need a large number of digits to determine the best performing competing scheme.

---

[1] http://ssli.ee.washington.edu/∼bilmes/uai06InferenceEvaluation/

Table 1: Results on various benchmarks. The columns Min, Avg, Per and Ord give the log-relative-error Δ for the minimum, the average, the martingale random permutation and the martingale order statistics heuristics respectively. The last two columns provide log-relative-error Δ and time for the bound propagation with cut-set conditioning scheme. In the first column N is the number of variables, D is the maximum domain size and E is the number of evidence variables. Time is in seconds. The column best LB reports the best lower bound reported by all competing scheme whose log-relative error is highlighted in each row.

|  |  | IJGP-sampling-Markov-LB | | | | | Bound Propagation | | Best |
| --- | --- | --- | --- | --- | --- | --- | --- | --- | --- |
|  | Exact | Min | Avg | Per | Ord |  | | | |
| $(N,D,|E|)$ | P(e) | Δ | Δ | Δ | Δ | Time | Δ | Time | LB |
| **Alarm** | | | | | | | | | |
| (100,2,36) | 2.8E-13 | 0.157 | **0.031** | 0.040 | 0.059 | 0.2 | 0.090 | 22.3 | 1.1E-13 |
| (100,2,51) | 3.6E-18 | 0.119 | **0.023** | 0.040 | 0.045 | 0.1 | 0.025 | 5.6 | 1.4E-18 |
| (125,2,55) | 1.8E-19 | 0.095 | **0.020** | 0.021 | 0.030 | 0.2 | 0.069 | 36.0 | 7.7E-20 |
| (125,2,71) | 4.3E-26 | 0.124 | **0.016** | 0.024 | 0.030 | 0.2 | 0.047 | 19.3 | 1.6E-26 |
| (125,2,46) | 8.0E-18 | 0.185 | **0.023** | 0.061 | 0.064 | 0.1 | 0.102 | 31.6 | 3.3E-18 |
| **CPCS** | | | | | | | | | |
| (360,2,20) | 1.3E-25 | 0.012 | 0.012 | **0.000** | 0.001 | 1.2 | 0.002 | 13.2 | 1.3E-25 |
| (360,2,30) | 7.6E-22 | 0.045 | 0.015 | 0.010 | 0.010 | 1.2 | **0.000** | 16.3 | 7.6E-22 |
| (360,2,40) | 1.2E-33 | 0.010 | 0.009 | 0.000 | **0.000** | 1.2 | **0.000** | 26.8 | 1.2E-33 |
| (360,2,50) | 3.4E-38 | 0.022 | 0.009 | 0.002 | **0.000** | 1.2 | **0.000** | 19.2 | 3.4E-38 |
| (422,2,20) | 7.2E-21 | 0.028 | 0.016 | 0.001 | 0.001 | 8.4 | **0.002** | 120 | 6.8E-21 |
| (422,2,30) | 2.7E-57 | 0.005 | 0.005 | 0.000 | **0.000** | 8.3 | **0.000** | 120 | 2.7E-57 |
| (422,2,40) | 6.9E-87 | 0.003 | 0.003 | 0.000 | **0.000** | 8.1 | 0.001 | 120 | 6.9E-87 |
| (422,2,50) | 1.4E-73 | 0.007 | 0.004 | 0.000 | **0.000** | 8.5 | 0.001 | 120 | 1.3E-73 |
| **Random** | | | | | | | | | |
| (53,50,6) | 4.0E-11 | 0.235 | 0.029 | 0.063 | **0.025** | 0.8 | 0.028 | 1.5 | 2.2E-11 |
| (54,50,5) | 2.1E-09 | 0.408 | 0.036 | 0.095 | **0.013** | 0.6 | 0.131 | 4.6 | 1.6E-09 |
| (57,50,6) | 1.9E-11 | 0.131 | 0.024 | **0.013** | 0.024 | 0.8 | 0.147 | 5.9 | 1.4E-11 |
| (58,50,8) | 1.6E-14 | 0.521 | **0.022** | 0.079 | 0.041 | 0.9 | 0.134 | 13.0 | 8.1E-15 |
| (76,50,15) | 1.5E-26 | 0.039 | 0.007 | **0.007** | 0.012 | 2.0 | 0.056 | 19.1 | 9.4E-27 |

|  |  | SampleSearch-Markov-LB | | | | | Bound Propagation | | |
| --- | --- | --- | --- | --- | --- | --- | --- | --- | --- |
|  | Exact | Min | Avg | Per | Ord |  | | | Best |
| $(N,D,|E|)$ | P(e) | Δ | Δ | Δ | Δ | Time | Δ | Time | LB |
| **Grid** | | | | | | | | | |
| (1156,2,120) | 9.1E-12 | 0.256 | **0.040** | 0.106 | 0.047 | 3.5 | 0.946 | 33.9 | 3.3E-12 |
| (1444,2,150) | 2.4E-12 | 0.208 | **0.094** | 0.111 | 0.107 | 5.3 | 3.937 | 600 | 2.0E-13 |
| (1444,2,150) | 3.5E-15 | 0.269 | **0.090** | 0.131 | 0.097 | 5.3 | 3.067 | 600 | 1.7E-16 |
| (1444,2,150) | 4.9E-10 | 0.243 | 0.093 | **0.090** | 0.108 | 4.3 | 5.380 | 600 | 7.2E-11 |
| (1444,2,150) | 4.6E-11 | 0.103 | 0.086 | **0.065** | 0.069 | 5.7 | 4.458 | 600 | 9.6E-12 |
| (1444,2,150) | 5.2E-14 | 0.127 | **0.100** | 0.079 | 0.098 | 3.7 | 3.456 | 600 | 4.6E-15 |
| **Linkage** | | | | | | | | | |
| (777,36,78) | 2.8E-54 | 0.243 | 0.176 | 0.169 | **0.153** | 8.6 | 1.022 | 3600 | 1.8E-62 |
| (2315,36,159) | 8.8E-72 | 0.390 | 0.340 | 0.347 | **0.340** | 40.7 | 1.729 | 3600 | 6.3E-96 |
| (1740,36,202) | 1.4E-111 | 0.438 | 0.235 | 0.323 | **0.235** | 30.8 | 0.984 | 3600 | 1.2E-137 |
| (2155,36,252) | 7.5E-151 | 0.196 | 0.128 | 0.196 | **0.128** | 41.6 | 0.298 | 3600 | 4.2E-170 |
| (2140,36,216) | 6.1E-114 | 0.419 | 0.311 | 0.354 | **0.311** | 48.3 | 1.560 | 3600 | 4.0E-149 |
| (749,36,66) | 2.2E-45 | 0.954 | 0.780 | 0.949 | **0.761** | 9.3 | 5.314 | 3600 | 2.4E-79 |
| (1820,36,155) | 2.1E-91 | 0.258 | 0.215 | 0.236 | **0.208** | 45.8 | 2.209 | 3600 | 3.0E-110 |
| (2155,36,169) | 1.4E-110 | 0.475 | 0.374 | 0.435 | **0.374** | 128.2 | 0.712 | 3600 | 1.2E-151 |
| (1020,36,135) | 2.8E-79 | 0.262 | 0.198 | 0.225 | **0.185** | 18.4 | 1.385 | 3600 | 8.9E-94 |
| **Two-layered** | | | | | | | | | |
| (1000,2,800) | 8.8E-26 | 0.059 | **0.024** | 0.032 | 0.029 | 12.2 | 11.01 | 3600 | 2.2E-26 |
| (1000,2,800) | 3.2E-28 | 0.076 | **0.030** | 0.045 | 0.042 | 10.5 | 9.95 | 3600 | 4.9E-29 |
| (1000,2,800) | 1.2E-27 | 0.061 | **0.019** | 0.020 | 0.024 | 7.7 | 10.19 | 3600 | 3.9E-28 |
| (1000,2,800) | 4.3E-26 | 0.109 | **0.050** | 0.060 | 0.061 | 13.0 | 10.87 | 3600 | 2.3E-27 |
| (1000,2,800) | 1.2E-26 | 0.115 | **0.036** | 0.055 | 0.046 | 20.5 | 10.61 | 3600 | 1.4E-27 |

## 5.2 Results

Our results are summarized in Table 1. We see that our new strategy of Markov-LB scales well with problem size and provides good quality high-confidence lower bounds on most problems. It clearly outperforms the bound propagation with cut-set conditioning scheme. We discuss the results in detail below.

**The Alarm networks** The Alarm networks are one of the earliest belief networks designed by medical experts for monitoring patients in intensive care. The evidence in these networks was set at random. These networks have between 100-125 binary nodes. We can see that Markov-LB is slightly superior to the bound propagation based scheme accuracy-wise, but is far more efficient time-wise. Among the different versions of Markov-LB, the average heuristic performs better than the martingale heuristics. The minimum heuristic is the worst performing heuristic.

**The CPCS networks** The CPCS networks are derived from the Computer-based Patient Case Simulation system [Pradhan et al., 1994]. The nodes of CPCS networks correspond to diseases and findings and conditional probabilities describe their correlations. The CPCS360b and CPCS422b networks have 360 and 422 variables respectively. We report results on the two networks with 20,30,40 and 50 randomly selected evidence nodes. We see that the lower bounds reported by the bound propagation based scheme are slightly better than Markov-LB on the CPCS360b networks but they take far more time. However, on the CPCS422b networks, Markov-LB has better performance than the bound propagation based scheme. The martingale heuristics (the random permutation and the order statistics) have better performance than the average heuristic. Again, the minimum heuristic has the worst performance. Note that we stop each algorithm after 2 mins of run-time if the algorithm has not terminated by itself.

**Random networks** The random networks are randomly generated graphs available from the UAI competition website. The evidence nodes are generated at random. The networks have between 53 and 76 nodes and the maximum domain size is 50. We see that Markov-LB is better than the bound propagation based scheme on all random networks. The random permutation and the ordered statistics martingale heuristics are slightly better than the average heuristic on most instances.

**Grid Networks** The Grid instances are also available from the UAI competition web-site. All nodes in the Grid are binary and evidence nodes are selected at random. The Grid networks have substantial amount of determinism and therefore we employ the SampleSearch based importance sampling scheme to compute the lower bound. Here, we stop each algorithm after 10 minutes if it has not terminated by itself. We notice that the performance of Markov-LB is significantly better than the bound propagation based scheme on all instances.

**Linkage networks** The linkage instances are generated by converting a Pedigree to a Bayesian network [Fishelson and Geiger, 2003]. These networks have be-tween 777-2315 nodes with a maximum domain size of 36 and are much larger than the Alarm, the CPCS, and the random networks. On these networks, we ran each algorithm until termination or until a time-bound of 1hr expired. The Linkage instances have a large number of zero probabilities which makes them hard for traditional importance sampling based schemes because of the rejection problem. Therefore, in all our experiments on linkage instances we used the SampleSearch based importance sampling scheme. On Linkage instances, IJGP-sampling did not return a single non-zero weight sample (not shown in Table 1) in more than one-hour of run-time yielding a lower bound of 0. We see that the bound propagation based scheme yields inferior lower bounds as compared to the SampleSearch based Markov-LB scheme. However, we notice that the log-relative-error is significantly higher for Markov-LB on the linkage instances than the Alarm, the CPCS, and the random instances. We suspect that this is because the quality of the proposal distribution computed from the output of IJGP is not as good on the linkage instances as compared to other instances. Finally, we notice that the average and the martingale-heuristics perform better than the min heuristic on all instances with the martingale order statistics heuristic being the best performing heuristic.

**Deterministic two-layered networks** Our final domain is that of completely deterministic two-layered networks. Here, the first layer is a set of root nodes connected to a second layer of leaf nodes. The CPTs of the root node are such that each value in the domain is equally likely while the CPTs associated with the leaf nodes are deterministic i.e. each CPT entry is either one or a zero. All nodes are binary. The evidence set is *all* the leaf nodes instantiated to a value. We experimented with 5 randomly generated 1000-variable two-layered networks each with 800 leaf nodes which are set to true (evidence). We employ the SampleSearch based importance sampling scheme for these networks because these instances have zero probabilities. We see that the average and the martingale heuristics are the best performing heuristics while the min-heuristic performs the worst. The SampleSearch based Markov-LB scheme shows significantly better performance than the bound propagation based scheme.

## 6 Conclusion and Summary

In this paper, we proposed a randomized approximation algorithm, *Markov-LB* for computing high confidence lower bounds on probability of evidence. Markov-LB is based on importance sampling and the Markov inequality. A straight forward application of the Markov inequality may lead to poor lower bounds and therefore we suggest various heuristic measures to improve Markov-LB's performance. We also show how the performance of Markov-LB can be improved further on belief networks with zero probabilities by using the SampleSearch scheme. Our experimental results on a range of benchmarks show that our new lower bounding scheme outperforms the state-of-the-art bound propagation scheme and provides high confidence good quality lower bounds on most instances.

### ACKNOWLEDGEMENTS

This work was supported in part by the NSF under award numbers IIS-0331707 and IIS-0412854.